\documentclass[runningheads]{llncs}
\usepackage[T1]{fontenc}
\usepackage{subcaption}
\usepackage{graphicx}
\usepackage{amsmath}
\usepackage{amsfonts}
\usepackage{amssymb}
\usepackage{enumitem}
\usepackage{booktabs}     
\usepackage{multirow}     
\usepackage{array}        

\usepackage{xcolor}       

\usepackage{hyperref}     

\usepackage{caption}      
\usepackage{marvosym}
\usepackage[textsize=tiny]{todonotes}

\begin{document}
\title{Multi-Agent Causal Reasoning for Suicide Ideation Detection Through Online Conversations}
\titlerunning{MACR}
%
\author{
Jun Li\inst{1}\and
Xiangmeng Wang\inst{1}\textsuperscript{(\Letter)}\and
Haoyang Li\inst{1}\and
Yifei Yan\inst{2}\and
Shijie Zhang\inst{3}\and
Hong Va Leong\inst{1}\and
Ling Feng\inst{4}\and
Nancy Xiaonan Yu\inst{2}\and
Qing Li\inst{1}
}
\authorrunning{J. Li et al.}

\institute{
The Hong Kong Polytechnic University, Hong Kong, China\\
\email{
hialex.li@connect.polyu.hk;\\
\{xiangmengpoly.wang, haoyang-comp.li, cshleong, qing-prof.li\}@polyu.edu.hk
}
\and
City University of Hong Kong, Hong Kong, China\\
\email{
yfyan8-c@my.cityu.edu.hk; xiaonayu@cityu.edu.hk
}
\and
Shenzhen MSU-BIT University, Shenzhen, China\\
\email{shijie.z@smbu.edu.cn}
\and
Tsinghua University, Beijing, China\\
\email{fengling@tsinghua.edu.cn}
}
%
\maketitle              
\begin{abstract}
Suicide remains a pressing global public health concern.  While social media platforms offer opportunities for early risk detection through online conversation trees, existing approaches face two major limitations: (1) They rely on predefined rules (e.g., quotes or relies) to log conversations that capture only a narrow spectrum of user interactions, and (2) They overlook hidden influences such as user conformity and suicide copycat behavior, which can significantly affect suicidal expression and propagation in online communities.
To address these limitations, we propose a Multi-Agent Causal Reasoning (MACR) framework that collaboratively employs a Reasoning Agent to scale user interactions and a Bias-aware Decision-Making Agent to mitigate harmful biases arising from hidden influences.
The Reasoning Agent integrates cognitive appraisal theory to generate counterfactual user reactions to posts, thereby scaling user interactions. 
It analyses these reactions through structured dimensions, i.e., cognitive, emotional, and behavioral patterns, with a dedicated sub-agent responsible for each dimension. 
The Bias-aware Decision-Making Agent mitigates hidden biases through a front-door adjustment strategy, leveraging the counterfactual user reactions produced by the Reasoning Agent. 
Through the collaboration of reasoning and bias-aware decision making, the proposed MACR framework not only alleviates hidden biases, but also enriches contextual information of user interactions with counterfactual knowledge. 
Extensive experiments on real-world conversational datasets demonstrate the effectiveness and robustness of MACR in identifying suicide risk.
\keywords{Multi-agent System \and Causal Inference \and Front-door adjustment  \and Suicide Risk Prediction}
\end{abstract}
\section{Introduction}

Suicide remains a critical global public health crisis.
Traditional clinical assessments face significant scalability challenges and are reactive in early identification of suicide risks~\cite{kessler2017predicting}. 
Social networks become increasingly popular and enable individuals with suicidal thoughts to share their mental struggles with a broad online audience. 
Thus, research has increasingly focused on analyzing online user interactions encompassing both posts and comments to enable proactive suicide risk prediction and to monitor the potential negative impacts of harmful content.




Analyzing online user interactions requires understanding their psychological and social dynamics. 
Psychological research~\cite{mesoudi2009cultural} shows that user interactions have temporal dependencies, where each interaction can influence subsequent content and tone. 
Sociological studies~\cite{avin2018elites} also reveal that response rates are highly uneven, with certain highly influential comments propagating quickly while some are left untouched. 
Most traditional suicide risk prediction methods fail to capture these dynamics. 
Early work~\cite{li2025dynaprotect,lee2023towards,sawhney2022towards,sawhney2021ordinal} typically models individual posts or comments using indicative factors such as linguistic features~\cite{lee2023towards}, psychological indicators~\cite{li2025dynaprotect}, and sentiment cues~\cite{sawhney2022towards}.
Such models, however, cannot effectively represent the sequential dependencies and dynamic emotional changes that occur across multiple interactions.

\begin{figure}
\centering
\includegraphics[width=\linewidth]{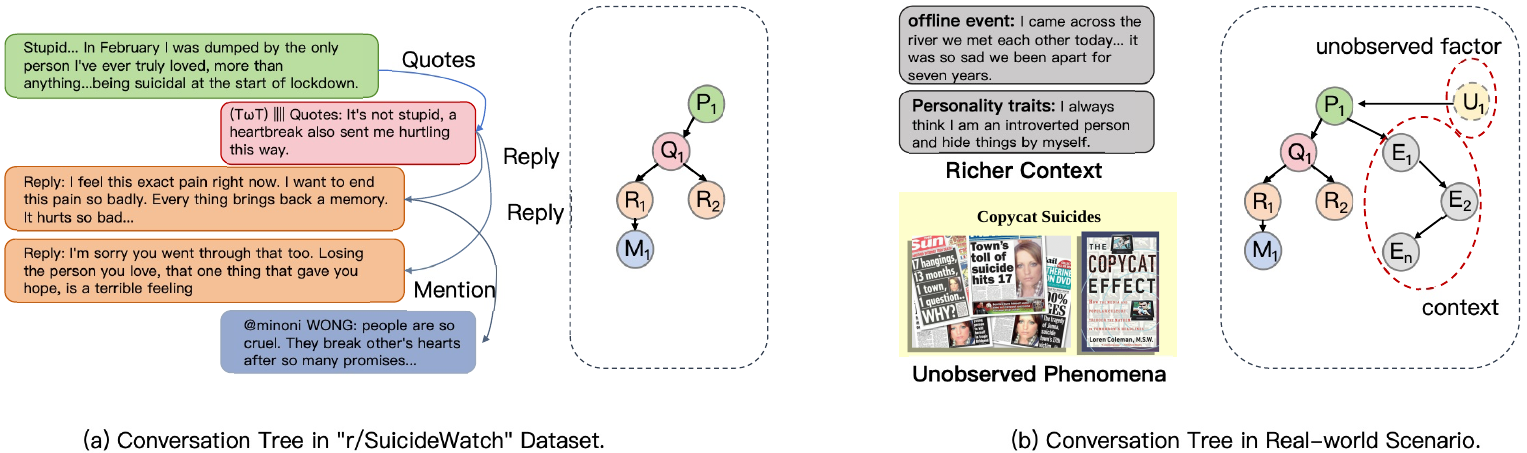}
\caption{(a) Example conversation tree from "r/SuicideWatch" in Reddit.(2) The limitations of current conversation tree-based methods}
\label{fig:Toy}
\vspace{-0.25in}
\end{figure}

Modeling user interactions as conversation trees~\cite{sawhney2022towards} emerges as a promising solution to capture unevenly distributed sequential user interactions. 
As shown in Figure~\ref{fig:Toy} (a), the conversation trees form hierarchical structures of user post-comments propagation, and exhibit scale-free characteristics, i.e., varying numbers of replies attached to different threads. 
Existing work has used conversation trees to model the effectiveness of comments and assess suicide risks on social media
For example, Kavuluru et al.~\cite{kavuluru2016classification} model the hierarchical conversation tree for automatic identification of helpful comments to assist moderators in providing timely support. Sawhney et al.~\cite{sawhney2022towards} explicitly modeled Reddit conversation trees to capture the scale-free dynamics and hierarchical branching patterns inherent in these tree structures for suicide ideation detection.
Other subsequent works~\cite{kumar2015detecting,boettcher2021studies} also use conversational trees for suicide ideation detection.

Despite these efforts, existing work on conversation tree-based suicide risk prediction methods suffers from significant limitations (cf. Figure~\ref{fig:Toy} (b)): 
1) They typically construct conversation trees based solely on explicitly logged interactions (e.g., mentions, quotes, or replies), capturing only a limited set of interactions~\cite{sawhney2021suicide}. However, user behavior is complex and influenced by various factors such as offline life events or personality traits, which are not reflected in logged interactions. 
By ignoring richer context, these methods provide an incomplete view of user behavior, reducing the accuracy of suicide risk detection.
(2) Their models assume the absence of unobserved factors~\cite{pearl2000causality}, an assumption that rarely holds in real-world contexts. Ignoring such unobserved factors introduces confounding bias, as models incorrectly treat observed content in the conversation tree as direct reflections of users' true psychological states. For example, sociological research~\cite{mesoudi2009cultural} has revealed the copycat suicide on social media, where exposure to suicidal content triggers imitative behaviors among vulnerable users. In such a phenomenon, an unobserved confounder like user conformity simultaneously influences what users post or comment (observed interactions) and their suicide risk. 
As such, by lacking richer context and ignoring unobserved influences, current models are limited in scalability and less robust at suicide ideation detection, especially in extreme or rapidly changing scenarios.

\noindent \textbf{Proposed Solution.} 
We propose to address incomplete contextual
and unobserved influences on user suicidality within online conversations from a novel causal view.
We propose \textbf{Multi-Agent Causal Reasoning (MACR)} that involves two major agents for collaboration: 1) a \textbf{Reasoning Agent} that scales the conversational tree by using four sub-agents to generate counterfactual user reactions to posts. 2) a \textbf{Bias-aware Decision-Making Agent} that controls the unobserved bias
with a causal adjustment strategy using generated counterfactual user reactions from the Reasoning Agent.
Our proposed Reasoning Agent
employs four language model-based agents that collaboratively simulate user reactions to posts, generating counterfactual reasoning nodes to capture potential yet unlogged user interactions.
The design of the four sub-agents and their workflow is inspired by the Paul-Elder critical thinking framework~\cite{elder2020critical}.
In addition, we innovatively integrate cognitive appraisal theory~\cite{lazarus1984stress} to guide the system in analyzing user reactions through structured dimensions, including cognitive, emotional, and behavioral patterns. 
The Bias-aware Decision-Making Agent adopts 
a causal method—especially front-door adjusted models—to mitigate the unobserved influence on user suicidality.
Front-door adjustment~\cite{pearl2009causality} uses an observable proxy mediator as the indicator of unobserved confounders to capture the causal pathway between treatment and outcome.
We construct the mediator variable as the conversation tree augmented with counterfactual reasoning nodes generated from our Reasoning Agent.
As a result, the proposed MACR framework mitigates unobserved confounding bias via front-door adjustment, while the mediator introduces additional counterfactual knowledge to enrich contextual information.
Our main contributions are summarized as follows:
\begin{itemize}
\item 
To the best of our knowledge, this work is the first to theoretically uncover the impact of unobserved confounders on user suicidality within online social conversations.

\item We innovatively integrate two complementary theoretical frameworks, the Paul-Elder critical thinking framework~\cite{elder2020critical} and 
cognitive 
appraisal theory~\cite{lazarus1984stress}, to guide the multi-agent reasoning process. 


\item We propose a confounding bias mitigation strategy based on front-door adjustment, which not only alleviates unobserved confounding bias but also enriches contextual information by incorporating counterfactual knowledge. 
\item Extensive experiments on real-world conversational datasets demonstrate the effectiveness and robustness of our framework in identifying suicide risk.
\end{itemize}

\begin{figure}
    \centering
    \includegraphics[width=0.75\linewidth]{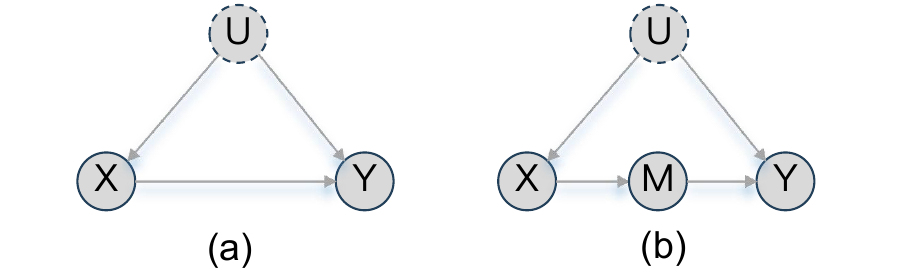}
    \caption{(a) The causality of conversation tree variable $X$ and suicide risk variable $Y$ is confounded by unobservable variable $U$. (b) The psychological inference generated by Reasoning Agent as a mediator variable $M$ between conversation tree variable $X$ and suicide risk variable $Y$.}
    \label{fig:scms}
\end{figure}

\section{Related Works}
\subsection{Suicide Risk Detection on Social Media}
Early research in suicide risk detection on social media adopted individual user modeling approaches, framing the problem as a binary or multi-class classification task at the user level. Initial efforts relied on feature engineering to extract meaningful patterns from user-generated content. Ji et al.~\cite{ji2018supervised} explored a comprehensive set of hand-crafted features across Twitter and Reddit platforms, subsequently testing their effectiveness with traditional classifiers and recurrent neural networks. Li et al.~\cite{li2021suicide} developed a hybrid architecture integrating BERT's contextual embeddings with bidirectional LSTM layers, enhancing the model through multi-task learning where emotion recognition served as an auxiliary objective. Similarly, Sawhney et al.~\cite{sawhney2021phase} introduced a temporal modeling framework that traces users' evolving emotional states through their posting history to better identify warning signs of suicidal ideation. Subsequent research began incorporating social contextual information. These approaches recognized that users' mental health states are influenced by their online social environments and peer interactions. Sawhney et al.~\cite{sawhney2021suicide} proposed combining individual emotional timelines with network-derived social signals. More recent work has turned to conversation tree structures as a richer representation of online discourse~\cite{sawhney2022towards,kavuluru2016classification}. Although these approaches successfully integrated social factors, their modeling of actual user interactions remained limited to platform-logged interactions such as explicit replies and mentions.


\section{Preliminaries}

\subsubsection{Structural Causal Model}
A Structural Causal Model (SCM)~\cite{pearl2016causal} provides a framework for causal relationships between variables in complex systems. In SCM, we typically use a directed acyclic graph termed the causal graph to represent the set of direct causal relationships between variables of interest.

\begin{definition}[Causal Graph]
\label{def:cg}
A \emph{Causal Graph}~\cite{bareinboim2022pearl} is a directed acyclic graph (DAG) $G=(\{\mathcal{V}, Z\}, \mathcal{E})$ represents causal relations among endogenous and exogenous variables.
$\mathcal{V}$ is a set of endogenous variables of interest, e.g., user comments.
$Z$ is a set of exogenous variables outside the model, e.g., user conformity.
$\mathcal{E}$ is edge set denoting causal relations among $G$.
\end{definition}

\begin{definition}[Confounder]
\label{def:cfounder}
Given a treatment variable and an outcome variable, a variable is considered a confounder~\cite{pearl2016causal} if it causally affects both the treatment and the outcome.
\end{definition}

\subsubsection{Front-door Adjustment}
Front-door adjustment estimates the causal effect of a treatment (or cause) on an outcome even when unobserved confounders exist. 
It works by introducing a mediator variable — an observable factor that lies on the causal pathway between the treatment and the outcome. By modeling how the treatment influences the mediator and how the mediator affects the outcome, front-door adjustment indirectly accounts for hidden confounding and enables unbiased causal estimation.

\section{Causal Analysis and Motivations}
\subsection{Causal Analysis}
Following Definition~\ref{def:cg}, we construct a causal graph as shown in Figure~\ref{fig:scms} (a) that models the causal relationships between variables in predicting suicide risk in user conversation trees. 
We explain the rationality of this casual graph here:
\begin{itemize}
    \item $X \to Y$: User conversation tree features ($X$) influence suicide risk ($Y$). Specifically, received comments such as expressions of hopelessness, negative emotional content, etc affect suicide risk.
    \item $U \to X$: Unobserved confounders (U) influence user conversation tree features (X). These latent factors include underlying mental health conditions, personality traits, life circumstances, and socioeconomic status that shape how users express themselves and interact on social media platforms.
    \item $U \to Y$: Unobserved confounders (U) directly influence suicide risk (Y). The same underlying factors that affect online conversation also contribute to the suicide risk.
\end{itemize}
Following Definition~\ref{def:cfounder}, $U$ is the confounder between $X$ and $Y$. Due to the existence of confounder $U$, we cannot estimate the intervened distribution of $P(Y=y \mid do(X=x))$ based on observed data distribution $P(Y=y \mid X=x)$, i.e., $P(Y=y \mid d o(X=x)) \neq P(Y=y \mid X=x)$. 
Instead, we need to account for and measuring all confounders' impact, i.e., $P(Y=y \mid \operatorname{do}(X=x))=\sum_{u} P(Y=y \mid X=x, U=u) P(U=u)$.

The above equation is based on the assumption that the confounder variable is measurable. However, in social media context, there may exist various unobserved or unmeasurable confounding variables. To address this setting, front-door adjustment offers a feasible solution.
Following the definition of front-door adjustment, 
we introduce a mediator $M$ as shown in Figure~\ref{fig:scms} (b).
Then, $P(Y|\text{do}(X))$ can be formulated as:

\begin{equation}
    P(Y|do(X)) = \sum_{m} P(m|do(X))P(Y|do(m))
\label{eq:frontdoor_1}
\end{equation}

The causal effect between $X$ and $Y$ is decomposed into two partially causal effects $P(m|\text{do}(X))$ and $P(Y|\text{do}(m))$.

To compute $P(Y|do(X))$, we need to block the backdoor path $X \leftarrow U \rightarrow Y \leftarrow M$. This path contains a collider structure $U \rightarrow Y \leftarrow M$. Therefore, the backdoor path is already blocked~\cite{pearl2016causal}, and we obtain:

\begin{equation}
P(m|do(X)) = P(m|X)
\label{eq:first_part_extension}
\end{equation}

To compute $P(Y|do(m))$, we must block confounding paths through $U$. The confounder $U$ affects both $Y$ (directly) and $M$ (through $X$). By conditioning on $X$, we intercept the confounding flow from $U$ to $M$. This blocks the backdoor path $M \leftarrow X \leftarrow U \rightarrow Y$, obtaining:
\begin{equation}
P(Y|do(m)) = \sum_{x \in X} P(x)P(Y|m, x)
\label{eq:second_part_extension}
\end{equation}

Finally, we obtain the 

\begin{equation}
    P(Y|do(X)) = \sum_{m} P(m|X)\sum_{x} P(x)P(Y|m, x)
\label{eq:frontdoor_2}
\end{equation}


\subsection{Motivations}
\label{sec:motivation}

\subsubsection{Agent-based Decomposition of Front-door Adjustment}
\label{sec:agent-decomposition}

Front-door adjustment naturally divides into two components, each implemented by a specialized agent. Intuitively, Equation~\eqref{eq:frontdoor_2} can be decomposed into two computational parts, corresponding to the distinct roles of the two agents:
\begin{equation}
    P(Y|do(X)) = \underbrace{\sum_{m} P(m|X)}_{Reasoning}\underbrace{\sum_{x} P(x)P(Y|m, x)}_{Decision-making}
\label{eq:frontdoor_2}
\end{equation}
We explain each computational part in the following.
\begin{itemize}

\item \textbf{Reasoning Agent.} This agent models the first part: $P(m|X)$. It shows how the conversation tree $X$ affects the mediator $M$. The agent generates counterfactual reasoning nodes using four sub-agents. These nodes simulate alternative user reactions to posts. By adding these counterfactual nodes to the original conversation tree, the agent creates an enriched mediator variable $m_i \in M$. This mediator captures both observed interactions and potential unlogged behaviors.

\item \textbf{Bias-aware Decision-making Agent.} This agent implements the decision-making component $\sum_{x} P(Y|m, x)P(x)$, predicting risk $Y$ from mediator $m_i \in M$. The summation over $x$ theoretically removes bias by averaging across all conversation variations. In practice, we approximate this intractable summation using an efficient sampling-based strategy. Details of this approximation are provided in Section~\ref{sec:decision_agent}.
\end{itemize}

\section{Method}\label{sec:method}

\begin{figure}
\centering
\includegraphics[width=1.0\linewidth]{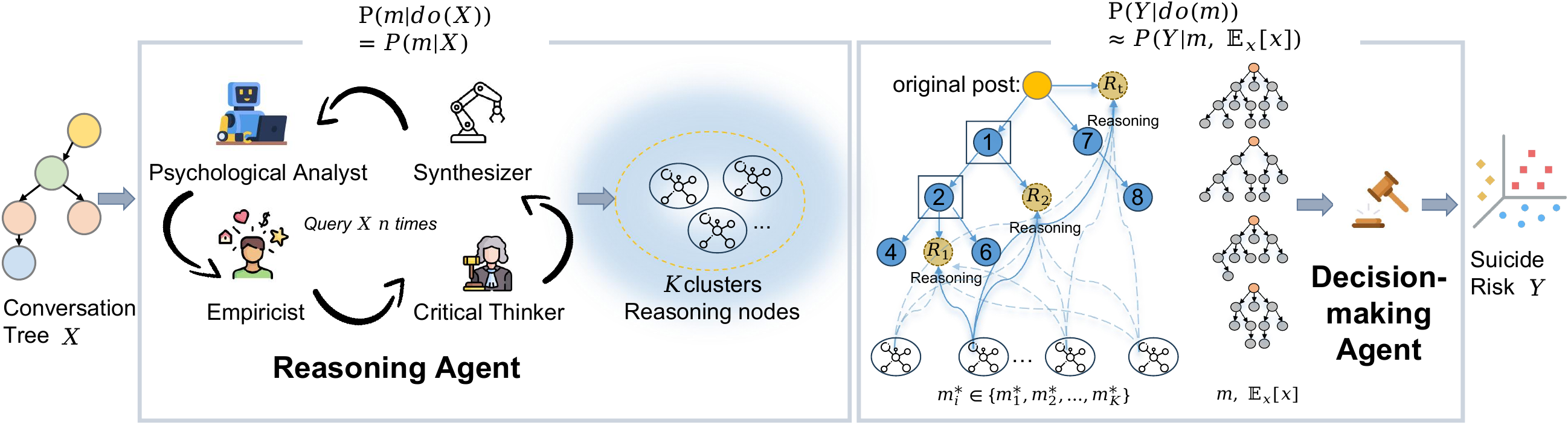}
\caption{Overview of Multi-Agent Causal
Reasoning (MACR) framework.}
\label{fig:overview_framework}
\end{figure}

Inspired by causal analysis and motivated by the need to decompose the front-door adjustment into two computational components aligned with distinct agents, we introduce the implementation of the proposed \textbf{Multi-Agent Causal Reasoning (MACR)} framework for robust and scalable suicide risk prediction. The overall architecture is illustrated in Figure~\ref{fig:overview_framework}.

\subsection{Reasoning Agent: Scalable Counterfactual Reaction Generator}\label{sec:reasoning_agent}

\noindent \textbf{Overview.} The Reasoning Agent generates counterfactual reasoning nodes to construct the mediator variable $M$. It consists of four collaborative sub-agents that simulate alternative user reactions.

\noindent \textbf{Architecture.} As shown in Figure~\ref{fig:overview_framework}, the agent processes input conversation tree $X$ through four 
sub-agents with different roles:
\begin{enumerate}
    \item \textit{Psychological Analyst}: This expert conducts in-depth psychological analysis of the original poster (OP), focusing on: Evaluating potential psychological reactions after receiving comments by analyzing the OP's psychological state, emotional patterns, and underlying motivations, coping styles and behavioral tendencies.
    \item \textit{Critical Thinker}: This expert identifies logical flaws and cognitive biases in the psychological analysis, including: Detecting cognitive biases (e.g., confirmation bias, availability heuristic). (2) Challenging logical validity of reasoning processes. (3) Identifying overlooked perspectives.
    \item \textit{Empiricist}: Grounding analysis in established psychological theories (e.g., Attachment Theory, Defense Mechanisms, Coping Strategies), this role functions as an "Evidence Police", focusing on empirical validation: (1) Identifying behavioral patterns based on textual evidence. (2) Rating evidence strength for each claim (Low/Medium/High). (3) Distinguishing between direct observations and speculative interpretations.
    \item \textit{Synthesizer}: The final reviewer who integrates all feedback and produces structured output: (1) Synthesizing insights from the Psychological Analyst, Critical Thinker, and Empiricist. (2) Filling analytical gaps and balancing depth with evidence constraints. (3) Generating comprehensive, balanced reasoning with clear evidence levels.
\end{enumerate}

\noindent \textbf{Workflow.} The system achieves collaboration through a three-round debate process: the first round is conducted by Psychological Analyst for preliminary analysis, the second round is questioned in parallel by Critical Thinker and Empiricist, and the third round is conducted by Synthesizer to integrate all feedback and generate the final output.

\noindent \textbf{Output.} Given a conversation tree $x$, the Raasonging Agent are queried to generates $n$ distinct counterfactual psychological inferences $M = \{m_1, m_2, \dots, m_n\}$, where each element $m_i$ represents a distinct counterfactual psychological inference. This adjustment facilitates exploration of diverse reasoning trajectories. Since multiple generated inferences may express semantically similar psychological states with different wordings, we consolidate them into representative nodes through clustering. Specifically, we first encode each inference $m_i$ into a dense semantic representation using the E5-large-v2 encoder~\cite{wang2022text}:

\begin{equation}
\bar{m}_i = \text{E5-Encoder}(m_i), \quad \bar{m}_i \in \mathbb{R}^{1024}
\label{eq:embedding}
\end{equation}

We then apply K-means clustering to group the $n$ embeddings $\{\bar{m}_1, \ldots, \bar{m}_n\}$ into $K$ clusters $\mathcal{C} = \{C_1, \ldots, C_K\}$ based on semantic similarity. For each cluster $C_i$, we select the inference closest to the cluster centroid as the representative node $m_i^*$:

\begin{equation}
m_i^* = \arg\min_{m_j \in C_i} |\bar{m}_j - \text{centroid}(C_i)|_2
\label{eq:representative}
\end{equation}

The probability of each representative inference is then estimated based on cluster size:

\begin{equation}
    P(m^*_i | do(x)) \approx \frac{|C_i|}{n}
\label{eq:cluster_probability}
\end{equation}

where $|C_i|$ denotes the number of inferences assigned to cluster $C_i$, and $n$ is the total number of generated counterfactual inferences.

\subsection{Bias-aware Decision-making Agent}\label{sec:decision_agent}

Based on the $K$ representative psychological impact assessments obtained through clustering in Section~\ref{sec:reasoning_agent}, we now estimate $P(Y|do(m))$ for each representative impact. For notational simplicity, we omit cluster indices in the following discussion.

$P(Y|do(m))$ measures the causal effect between the inferred psychological impacts $m$ and the subsequent suicide risk level $Y$. As established in Eq.~\eqref{eq:second_part_extension}: 
\begin{align}
P(Y|\text{do}(m)) &= \sum_{x \in \mathcal{X}} P(x)P(Y|m,x) \nonumber\\
&= \mathbb{E}_{x}[P(Y|m,x)]
\label{eq:fd_formula_expansion}
\end{align}
where $x \in \mathcal{X}$ represents specific instances of post-comment trees in the input space, and $m$ is the inferred psychological impact.

\noindent \textbf{Approximating the Expectation.} Since enumerating all possible post-comment trees $x \in \mathcal{X}$ is intractable, we employ a two-stage approximation strategy.

First, following the normalized weighted geometric mean (NWGM) approximation framework~\cite{chen2023causal,tian2022debiasing}, under mild smoothness assumptions on the Decision-Making Agent (DA), the expectation over the input distribution can be approximated by:
\begin{equation}
\mathbb{E}_{x}[P(Y|m,x)] \approx P(Y|m,\mathbb{E}_{x}[x])
\label{eq:nwgm_approximation}
\end{equation}

However, directly computing $\mathbb{E}_x[x]$ is infeasible for discrete structured data like post-comment trees. To address this challenge, we approximate the expectation by integrating two sources of information: (1) the \textit{input comment tree} $x$ used by the Reasoning Agent to generate psychological impacts $M$, where we augment $x$ by attaching the inferred impacts $m_i \in M$ as additional nodes, and (2) the \textit{broader input distribution} captured through stratified demonstrations to debias the prediction. We achieve this through \textit{in-context learning with stratified demonstrations}.

Specifically, the Decision-Making Agent receives the augmented instance $(x, m_i)$ along with stratified demonstrations $D$ to approximate $P(Y|m,\mathbb{E}_{x}[x])$. The demonstrations provide contextual information about $P(x)$ across different risk levels. For each input $x$, we construct $D_i$ by selecting one representative example from each of the four risk levels based on Cosine semantic similarity between them and $x$. This stratified sampling ensures distributional coverage across risk levels while retrieving contextually relevant examples.

\noindent \textbf{Output.} The Decision-Making Agent is implemented as a prompted large language model. Given the input $x$ and $m_i^*$ and matching demonstrations, it predicts:
\begin{align}
P(Y|\text{do}(m_i^*)) &\approx P(Y|m_i^*,\mathbb{E}_{x}[x]) \\ 
&\approx \text{DA}(m_i^*, x, {D_i})
\label{eq:da_with_demos}
\end{align}

Here, $D_i$ plays a critical role in approximating $\mathbb{E}_{x}[x]$: by exposing the agent to stratified examples from all risk levels, the demonstrations provide representative instances that help approximate the expectation over $X$, thereby debiasing the prediction. For each $m_i^*$, we perform $T$ independent predictions using the Decision-Making Agent and aggregate the predictions via majority voting:

\begin{equation}
P(Y| do(m_i^*)) \approx \frac{1}{T} \sum_{t=1}^{T} \mathbb{I}[Y=y_{t}], \quad y_t = \text{DA}_t(m_{i}^*, x, D_i)
\label{eq:voting}
\end{equation}

where $\text{y}_t$ denotes the $t$-th prediction of the Decision-Making Agent, and $\mathbb{I}[\cdot]$ is the indicator function.


\noindent \textbf{Final Risk Prediction.} Based on the above Eq.~\eqref{eq:cluster_probability} and Eq.~\eqref{eq:voting}, we combine the estimates from Reasonging Agent and Decision-making Agent, the complete front-door adjustment becomes:
\begin{align}
P(Y=y|do(X)) &= \sum_{m} P(m|do(X))P(Y|do(m)) \nonumber \\
&=\sum_{i=1}^{K} \frac{|C_i|}{n} \cdot \frac{1}{T} \sum_{t=1}^{T} \mathbb{I}[Y=\text{DA}_t(m_i^*, x, D_i)]
\label{eq:final_frontdoor}
\end{align}

\section{Experiments}

\subsection{Experimental Dataset}
We conduct experiments on two datasets to evaluate our framework: (1) the Suicidal Comment Tree Dataset~\cite{li2025suicidal}, and (2) Protective Factor-Aware Dataset with Comment Tree constructed by augmenting posts in Protective Factor-Aware dataset~\cite{li2025protective} with associated comment trees.

\subsection{Baselines}
We compare our method against four categories of baselines: 1) \textbf{Suicide Risk Prediction Models}: including \textbf{DAFIL}~\cite{li2025dynaprotect}, \textbf{TSAML}~\cite{lee2023towards}, \textbf{STATENet}~\cite{sawhney2020time} and \textbf{SISMO}~\cite{sawhney2021ordinal}; 2) \textbf{Debiased Language Models}:  We implement back-door adjustment~\cite{chen2023causal} on Qwen-3-4B-Instruct to mitigate reliance on psycholinguistic and affective shortcuts, result in \textbf{PLM-LIWC} that treats 18 LIWC psychological categories~\cite{pennebaker2015development} as confounders for causal debiasing; and \textbf{PLM-Emotion} that uses Plutchik's eight primary emotions~\cite{plutchik1980general} as confounders for demonstration retrieval.
3) \textbf{Large Language Models}: We evaluate Llama-3-8B-Instruct, DeepSeek-Chat, and GPT-5 as strong foundation model baselines. 4) \textbf{Graph Neural Network Models}: We adapt classic GNN architectures including HCN~\cite{sawhney2022towards}, GCN~\cite{kipf2016semi}, GraphSAGE~\cite{hamilton2017inductive}, and GAT~\cite{velivckovic2017graph} to post-comment tree structures.

\subsection{Experimental Setup}
In our multi-agent framework, we employ Qwen 3-4B-Instruct for the three specialized agents within the Reasoning Agent (Psychological Analyst, Empiricist, and Critical Thinker), while DeepSeek-Chat powers both the Synthesizer in the Reasoning Agent and the Decision-Making Agent. We set the number of reasoning iterations to $n=10$, the number of clusters to $K=3$. For the Decision-Making Agent, we perform $T=3$ independent predictions and aggregate them via Eq.~\eqref{eq:final_frontdoor} for final prediction. Following standard practice in multi-class classification for suicide risk prediction, we evaluate our framework using Weighted-Precision, Weighted-Recall, and Weighted-F1 as our primary evaluation metrics. 

\subsection{Overall Comparison Results}

\begin{table*}[t]
\centering
\caption{Overall performance comparison on two datasets for suicide risk prediction. Best results are in \textbf{bold}, second best are \underline{underlined}. All values are mean ± std over 5 folds. PFA: Protective Factor-Aware Dataset; SCT: Suicidal Comment Tree Dataset.}
\label{tab:main_results}
\resizebox{\textwidth}{!}{
\begin{tabular}{lcccccc}
\toprule
\multirow{2}{*}{\textbf{Model}} & \multicolumn{3}{c}{\textbf{PFA Dataset}} & \multicolumn{3}{c}{\textbf{SCT Dataset}} \\
\cmidrule(lr){2-4} \cmidrule(lr){5-7}
& \textbf{W-Prec.} & \textbf{W-Rec.} & \textbf{W-F1} & \textbf{W-Prec.} & \textbf{W-Rec.} & \textbf{W-F1} \\
\midrule
\multicolumn{7}{l}{\textit{Category 1: Suicide Risk Prediction Models}} \\
DAFIL & 0.3487 ± 0.0180 & 0.3519 ± 0.0215 & \underline{0.3433 ± 0.0201} & 0.4197 ± 0.0448 & 0.4151 ± 0.0319 & 0.4103 ± 0.0391 \\
TSAML & 0.2773 ± 0.0386 & \textbf{0.3998 ± 0.0279} & 0.3083 ± 0.0354 & 0.2871 ± 0.0932 & 0.3769 ± 0.0337 & 0.2704 ± 0.0704 \\
STATENet & 0.2346 ± 0.0623 & 0.3046 ± 0.0676 & 0.2250 ± 0.0461 & 0.2308 ± 0.0736 & 0.2939 ± 0.0961 & 0.2238 ± 0.0933 \\
SISMO & 0.3092 ± 0.0269 & 0.3172 ± 0.0141 & 0.3049 ± 0.0159 & 0.2722 ± 0.0829 & 0.3902 ± 0.0317 & 0.3036 ± 0.0695 \\
\midrule
\multicolumn{7}{l}{\textit{Category 2: Debiased Language Models}} \\
PLM-LIWC & 0.3739 ± 0.0351 & 0.3744 ± 0.0283 & 0.3262 ± 0.0257 & 0.5368 ± 0.0599 & 0.4764 ± 0.0366 & 0.4345 ± 0.0397 \\
PLM-Emotion & 0.3627 ± 0.0796 & 0.3608 ± 0.0301 & 0.3110 ± 0.0357 & \underline{0.5600 ± 0.0325} & \underline{0.4873 ± 0.0412} & \underline{0.4480 ± 0.0424} \\
\midrule
\multicolumn{7}{l}{\textit{Category 3: LLMs}} \\
Llama-3-8B-Instruct & 0.3225 ± 0.0110 & 0.3420 ± 0.0092 & 0.3306 ± 0.0090 & 0.4447 ± 0.0866 & 0.3743 ± 0.0256 & 0.3723 ± 0.0321 \\
DeepSeek-Chat & \underline{0.4237 ± 0.0290} & 0.3535 ± 0.0266 & 0.3431 ± 0.0206 & {0.4312 $\pm$ 0.0246} & {0.3669 $\pm$ 0.0239} & {0.3438 $\pm$ 0.0196} \\
GPT-5 & 0.1748 ± 0.0676 & 0.3738 ± 0.0311 & 0.2130 ± 0.0311 & 0.5445 ± 0.0414 & 0.3042 ± 0.0399 & 0.3316 ± 0.0257 \\
\midrule
\multicolumn{7}{l}{\textit{Category 4: Graph Neural Networks}} \\
HCN & 0.3025 ± 0.0522 & 0.2613 ± 0.0457 & 0.2568 ± 0.0440 & 0.4356 ± 0.1263 & 0.3320 ± 0.0930 & 0.3282 ± 0.0877 \\
GCN & 0.3515 ± 0.0515 & 0.3632 ± 0.0355 & 0.3313 ± 0.0329 & 0.4471 ± 0.0667 & 0.4791 ± 0.0587 & 0.4320 ± 0.0691 \\
GraphSAGE & 0.3460 ± 0.0538 & 0.3695 ± 0.0296 & 0.3409 ± 0.0423 & 0.4129 ± 0.0535 & 0.4572 ± 0.0378 & 0.3995 ± 0.0401 \\
GAT & 0.3549 ± 0.0465 & \underline{0.3852 ± 0.0422} & 0.3425 ± 0.0442 & 0.4268 ± 0.0690 & 0.4745 ± 0.0529 & 0.4187 ± 0.0574 \\
\midrule
\textbf{Ours} & \textbf{0.4269 $\pm$ 0.0467} & {0.3765 $\pm$ 0.0365} & \textbf{0.3768 $\pm$ 0.0373} & \textbf{0.5629 $\pm$ 0.0554} & \textbf{0.5189 $\pm$ 0.0268} & \textbf{0.5108 $\pm$ 0.0345} \\
\bottomrule
\end{tabular}
}
\vspace{-0.2in}
\end{table*}

Table~\ref{tab:main_results} presents the overall performance comparison between our proposed framework and baseline methods across both datasets. Our method consistently achieves the best performance on the majority of metrics, demonstrating the effectiveness of causally-principled reasoning for suicide risk prediction. On the PFA dataset, our approach achieves a Weighted-F1 of 0.3768, outperforming the best baseline (DAFIL) by 8.9\%. The improvement is even more pronounced on the SCT dataset, where our method achieves 0.5108 Weighted-F1, representing a 12.3\% gain over the second-best model (PLM-Emotion). Notably, debiased language models (PLM-LIWC, PLM-Emotion) show competitive performance but still fall short of our method, suggesting that lexical-level back-door adjustment alone is insufficient for capturing the complex psychological mechanisms in social interactions. Similarly, while GNN-based models (HCN, GCN, GraphSAGE, GAT) effectively capture structural information from comment trees, they lack the explicit causal reasoning pathway necessary to model how social interactions mediate the relationship between user context and suicide risk. These results validate that our front-door adjustment framework, by explicitly decomposing causal effects through psychological impact mediators, provides a more principled and effective approach to suicide risk prediction.

Table~\ref{tab:main_results} presents the overall performance comparison between our proposed front-door adjustment framework and baseline methods across both datasets. Our method demonstrates superior performance, achieving the best Weighted-F1 scores on both datasets while maintaining competitive precision and recall. On the PFA dataset, our approach achieves a Weighted-F1 of 0.3768, outperforming the best baseline (DAFIL) by 8.9\%. The improvement is even more pronounced on the SCT dataset, where our method achieves 0.5108 Weighted-F1, representing a 12.3\% gain over the second-best model (PLM-Emotion).

The superior performance of our framework can be attributed to two key advantages. First, our method addresses unobserved confounding more effectively than back-door adjustment approaches. While debiased language models (PLM-LIWC, PLM-Emotion) attempt to mitigate bias through back-door adjustment by conditioning on predefined psycholinguistic features and emotions, this approach has fundamental limitations: (1) it requires exhaustive measurement of all confounders, which is impractical in complex social contexts where factors like user conformity, copycat behavior, and offline life events remain unobservable; (2) it operates at the lexical level, capturing only surface-level linguistic patterns rather than the underlying psychological mechanisms that drive suicide risk. In contrast, our front-door adjustment circumvents the need to measure unobserved confounders by introducing psychological impact assessments as mediators, explicitly modeling the causal pathway $X \rightarrow M \rightarrow Y$ and thereby providing theoretically sound deconfounding even when $U$ remains unmeasured.

Second, our multi-agent reasoning framework enriches contextual information beyond what is available in logged interactions. Traditional methods, including both GNN-based models (HCN, GCN, GraphSAGE, GAT) and sequential models (DAFIL, TSAML, STATENet, SISMO), are constrained to analyzing only explicitly logged interactions such as replies and mentions. While GNN models effectively capture structural patterns in conversation trees and sequential models model temporal dependencies, both categories suffer from incomplete context. Our Reasoning Agent addresses this limitation by generating counterfactual reasoning nodes that simulate alternative user reactions grounded in cognitive appraisal theory. This augmentation captures a richer spectrum of possible psychological responses, providing the Decision-Making Agent with more comprehensive contextual information for risk assessment.

Third, our framework provides structured causal reasoning that outperforms generic LLM prompting. While large language models (Llama-3-8B-Instruct, DeepSeek-Chat, GPT-5) demonstrate strong general-purpose capabilities, they lack explicit causal mechanisms to handle confounding bias in suicide risk prediction. These models rely on pattern matching from their training data, making them susceptible to spurious correlations between conversation features and risk levels. Our multi-agent collaboration, guided by the Paul-Elder critical thinking framework and cognitive appraisal theory, systematically decomposes the prediction task into causally principled components: the Reasoning Agent generates theory-grounded psychological mediators, while the Decision-Making Agent performs front-door adjustment to block confounding paths. This structured approach ensures predictions are based on genuine causal mechanisms rather than superficial linguistic patterns, explaining why our method achieves substantial improvements over even the most advanced LLMs (e.g., 9.8\% higher Weighted-F1 than DeepSeek-Chat on PFA dataset).



\subsection{Ablation Study}

To validate the contribution of each component in our Multi-Agent Causal Reasoning (MACR) framework, we conduct comprehensive ablation experiments. Results are shown in Tables~\ref{tab:ablation}.

\textbf{w/o Reasoning Agent}: This variant removes the first causal component $P(M|do(X))$ entirely. Instead of generating counterfactual reasoning nodes through multi-agent collaboration, we directly feed the original conversation tree $X$ to the Decision-Making Agent for risk prediction: $P(Y|do(X)) \approx DA(X)$.

\textbf{w/o Decision-Making Agent}: This variant removes the second causal component $P(Y|do(M))$. We generate and cluster counterfactual reasoning nodes to obtain $K$ representative nodes $M = \{m_1^*, m_2^*, \cdots, m_K^*\}$ with probabilities $P(m_k^*|X) = \frac{|C_k|}{n}$. For each $m_k^*$, we directly prompt the Qwen-3-4B-Instruct to predict risk: $\hat{y}_k = \text{LLM}(Y|m_k^*)$, and aggregate predictions through weighted majority voting.



\textbf{w/o Multi-Agent Collaboration}: This variant uses only the Psychological Analyst to generate counterfactual reasoning nodes, removing the collaborative refinement from Critical Thinker, Empiricist, and Synthesizer. This examines whether multi-agent collaboration is essential for effective mediators.

The ablation study results in Table~\ref{tab:ablation} demonstrate the importance of each component in our MACR framework. Removing the Reasoning Agent leads to the most substantial performance degradation, with Weighted-F1 dropping by 32.7\%, validating that counterfactual reasoning nodes are crucial for enriching contextual information and enabling effective causal deconfounding. The w/o Decision-Making Agent variant shows a 7.8\% decline in Weighted-F1, indicating that the front-door adjustment mechanism P(Y|do(M)) is essential for mitigating confounding bias through proper causal conditioning. Notably, the w/o Multi-Agent Collaboration variant, which relies solely on the Psychological Analyst without collaborative refinement, achieves only 3.2\% lower performance than the full model. This relatively small gap suggests that while multi-agent collaboration provides marginal improvements, the primary benefits stem from combining counterfactual augmentation with causal adjustment. Overall, these findings confirm that both the reasoning and decision-making components are essential for robust suicide risk prediction.


\begin{table}[h]
\centering
\caption{Ablation study on the SCT dataset. We systematically remove or modify key components to examine their contributions.}
\label{tab:ablation}
\resizebox{0.9\textwidth}{!}{
\begin{tabular}{lccc}
\toprule
\textbf{Model Variant} & \textbf{Weighted-Precision} & \textbf{Weighted-Recall} & \textbf{Weighted-F1} \\
\midrule
\textbf{Full Model (Ours)} & \textbf{{0.5629 $\pm$ 0.0554}} & \textbf{{0.5189 $\pm$ 0.0268}} & \textbf{{0.5108 $\pm$ 0.0345}} \\
\midrule
w/o Reasoning Agent & {0.4312 $\pm$ 0.0246} & {0.3669 $\pm$ 0.0239} & {0.3438 $\pm$ 0.0196} \\
 & {($-$23.4\%)} & {($-$29.3\%)} & {($-$32.7\%)} \\
w/o Decision-Making Agent & {0.5492 $\pm$ 0.0466} & {0.4884 $\pm$ 0.0289} & {0.4708 $\pm$ 0.0260} \\
 & {($-$2.4\%)} & {($-$5.9\%)} & {($-$7.8\%)} \\
w/o Multi-Agent Collaboration & {0.5534 $\pm$ 0.0650} & {0.5067 $\pm$ 0.0322} & {0.4942 $\pm$ 0.0390} \\
 & {($-$1.7\%)} & {($-$2.4\%)} & {($-$3.2\%)} \\
\bottomrule
\end{tabular}
}
\vspace{-0.2in}
\end{table}

\subsection{Analysis of Prompting Techniques}

Comparison of our MACR framework against conventional prompting strategies on the PFA dataset.

\begin{table}[h]
\centering
\caption{Comparison with different prompting techniques. All methods use the same base model and comment tree information.}
\label{tab:prompting}
\resizebox{0.9\textwidth}{!}{
\begin{tabular}{lccc}
\toprule
\textbf{Method} & \textbf{Weighted-Precision} & \textbf{Weighted-Recall} & \textbf{Weighted-F1} \\
\midrule
In-Context Learning (ICL) & {0.4064} & {0.3457} & {0.3247} \\
Chain-of-Thought (CoT) & {0.4077} & {0.3457} & {0.3201} \\
\midrule
\textbf{Ours (Front-door)} & {\textbf{0.4269}} & {\textbf{0.3765}} & {\textbf{0.3768}} \\
\bottomrule
\end{tabular}
}
\vspace{-0.2in}
\end{table}

To demonstrate the effectiveness of our front-door adjustment framework, we compare it against conventional prompting strategies on the PFA dataset, as shown in Table~\ref{tab:prompting}. All methods use the same base model and comment tree information to ensure fair comparison. Our front-door adjustment approach achieves the best performance across all metrics (Weighted-F1: 0.3768), outperforming both In-Context Learning (ICL) and Chain-of-Thought (CoT) prompting. While ICL provides demonstrations to guide the model, it lacks the explicit causal decomposition $P(Y|do(X)) = \sum_m P(m|X)\sum_xP(x)P(Y|m,x)$, resulting in 13.8\% lower performance compared to our method. Similarly, although CoT generates intermediate reasoning steps, it does not explicitly model psychological impacts as mediators or perform proper causal adjustment by conditioning on both mediators and conversation tree $X$, leading to a 15.0\% performance gap. These results validate that our structured front-door adjustment provides a principled framework for handling confounding biases in suicide risk prediction, significantly outperforming generic prompting strategies.

\section{Ethics Statement}
We take the following ethical precautions: (1) All data is anonymized with identifying user's information removed; (2) This system is designed as a screening tool to assist human experts, not for autonomous decision-making; (3) We acknowledge the potential for false negatives/positives and emphasize the need for professional clinical validation; (4) The system should not be deployed without human oversight and appropriate crisis intervention protocols.

\section{Conclusion}
In this paper, we propose the Multi-Agent Causal Reasoning (MACR) framework to address two critical limitations in existing suicide risk prediction methods: the narrow scope of logged interactions and unobserved confounders. Our framework consists of two key components. The Reasoning Agent enriches conversation trees with psychologically-grounded counterfactual user reactions. The Bias-aware Decision-Making Agent applies front-door adjustment to mitigate unobserved confounding bias. Extensive experiments on real-world datasets demonstrate that MACR achieves state-of-the-art performance through structured causal reasoning. Future work will extend this multi-agent causal paradigm to other mental health monitoring tasks and refine the cognitive appraisal process for finer-grained risk assessment.

\bibliographystyle{splncs04}
\bibliography{bibs}

@book{pearl2016causal, 
    title={Causal inference in statistics: A primer}, 
    author={Pearl, Judea and Glymour, Madelyn and Jewell, Nicholas P}, year={2016}, 
    publisher={John Wiley \& Sons} 
}

@article{kessler2017predicting,
  title={Predicting suicides after outpatient mental health visits in the Army Study to Assess Risk and Resilience in Servicemembers (Army STARRS)},
  author={Kessler, Ronald C and Stein, Murray B and Petukhova, Maria V and Bliese, Paul and Bossarte, Robert M and Bromet, Evelyn J and Fullerton, Carol S and Gilman, Stephen E and Ivany, Christopher and Lewandowski-Romps, Lisa and others},
  journal={Molecular psychiatry},
  volume={22},
  number={4},
  pages={544--551},
  year={2017},
  publisher={Nature Publishing Group}
}

@article{avin2018elites,
  title={Elites in social networks: An axiomatic approach to power balance and Price’s square root law},
  author={Avin, Chen and Lotker, Zvi and Peleg, David and Pignolet, Yvonne-Anne and Turkel, Itzik},
  journal={PloS one},
  volume={13},
  number={10},
  pages={e0205820},
  year={2018},
  publisher={Public Library of Science San Francisco, CA USA}
}

@article{mesoudi2009cultural,
  title={The cultural dynamics of copycat suicide},
  author={Mesoudi, Alex},
  journal={PLoS One},
  volume={4},
  number={9},
  pages={e7252},
  year={2009},
  publisher={Public Library of Science San Francisco, USA}
}

@inproceedings{kavuluru2016classification,
  title={Classification of helpful comments on online suicide watch forums},
  author={Kavuluru, Ramakanth and Ramos-Morales, Mar{\'\i}a and Holaday, Tara and Williams, Amanda G and Haye, Laura and Cerel, Julie},
  booktitle={Proceedings of the 7th ACM international conference on bioinformatics, computational biology, and health informatics},
  pages={32--40},
  year={2016}
}

@book{pearl2009causality,
  title={Causality},
  author={Pearl, Judea},
  year={2009},
  publisher={Cambridge university press}
}

@article{boettcher2021studies,
  title={Studies of depression and anxiety using reddit as a data source: scoping review},
  author={Boettcher, Nick},
  journal={JMIR mental health},
  volume={8},
  number={11},
  pages={e29487},
  year={2021},
  publisher={JMIR Publications Toronto, Canada}
}

@book{elder2020critical,
  title={Critical thinking: Tools for taking charge of your learning and your life},
  author={Elder, Linda and Paul, Richard},
  year={2020},
  publisher={Rowman \& Littlefield}
}

@inproceedings{kumar2015detecting,
  title={Detecting changes in suicide content manifested in social media following celebrity suicides},
  author={Kumar, Mrinal and Dredze, Mark and Coppersmith, Glen and De Choudhury, Munmun},
  booktitle={Proceedings of the 26th ACM conference on Hypertext \& Social Media},
  pages={85--94},
  year={2015}
}

@book{lazarus1984stress,
  title={Stress, appraisal, and coping},
  author={Lazarus, Richard S},
  volume={445},
  year={1984},
  publisher={Springer}
}

@article{wang2022text,
  title={Text embeddings by weakly-supervised contrastive pre-training},
  author={Wang, Liang and Yang, Nan and Huang, Xiaolong and Jiao, Binxing and Yang, Linjun and Jiang, Daxin and Majumder, Rangan and Wei, Furu},
  journal={arXiv preprint arXiv:2212.03533},
  year={2022}
}

@article{li2025suicidal,
  title={Suicidal Comment Tree Dataset: Enhancing Risk Assessment and Prediction Through Contextual Analysis},
  author={Li, Jun and Zhao, Qun},
  journal={arXiv preprint arXiv:2510.14395},
  year={2025}
}

@incollection{bareinboim2022pearl,
  title={On Pearl’s hierarchy and the foundations of causal inference},
  author={Bareinboim, Elias and Correa, Juan D and Ibeling, Duligur and Icard, Thomas},
  booktitle={Probabilistic and causal inference: the works of judea pearl},
  pages={507--556},
  year={2022}
}

@inproceedings{li2021suicide,
  title={Suicide ideation detection on social media during COVID-19 via adversarial and multi-task Learning},
  author={Li, Jun and Yan, Zhihan and Lin, Zehang and Liu, Xingyun and Leong, Hong Va and Yu, Nancy Xiaonan and Li, Qing},
  booktitle={Asia-Pacific Web (APWeb) and Web-Age Information Management (WAIM) Joint International Conference on Web and Big Data},
  pages={140--145},
  year={2021},
  organization={Springer}
}

@article{ji2018supervised,
  title={Supervised learning for suicidal ideation detection in online user content},
  author={Ji, Shaoxiong and Yu, Celina Ping and Fung, Sai-fu and Pan, Shirui and Long, Guodong},
  journal={Complexity},
  volume={2018},
  number={1},
  pages={6157249},
  year={2018},
  publisher={Wiley Online Library}
}

@inproceedings{sawhney2022towards,
  title={Towards suicide ideation detection through online conversational context},
  author={Sawhney, Ramit and Agarwal, Shivam and Neerkaje, Atula Tejaswi and Aletras, Nikolaos and Nakov, Preslav and Flek, Lucie},
  booktitle={Proceedings of the 45th international ACM SIGIR conference on research and development in information retrieval},
  pages={1716--1727},
  year={2022}
}

@inproceedings{sawhney2021suicide,
  title={Suicide ideation detection via social and temporal user representations using hyperbolic learning},
  author={Sawhney, Ramit and Joshi, Harshit and Shah, Rajiv and Flek, Lucie},
  booktitle={Proceedings of the 2021 conference of the North American Chapter of the Association for Computational Linguistics: human language technologies},
  pages={2176--2190},
  year={2021}
}

@misc{pearl2000causality,
  title={Causality: Models, Reasoning, and Inference/Cambridge Univ},
  author={Pearl, J},
  year={2000},
  publisher={Press}
}

@inproceedings{chen2023causal,
  title={Causal intervention and counterfactual reasoning for multi-modal fake news detection},
  author={Chen, Ziwei and Hu, Linmei and Li, Weixin and Shao, Yingxia and Nie, Liqiang},
  booktitle={Proceedings of the 61st Annual Meeting of the Association for Computational Linguistics (Volume 1: Long Papers)},
  pages={627--638},
  year={2023}
}

@inproceedings{tian2022debiasing,
  title={Debiasing nlu models via causal intervention and counterfactual reasoning},
  author={Tian, Bing and Cao, Yixin and Zhang, Yong and Xing, Chunxiao},
  booktitle={Proceedings of the AAAI Conference on Artificial Intelligence},
  volume={36},
  number={10},
  pages={11376--11384},
  year={2022}
}

@inproceedings{li2025dynaprotect,
  title={DynaProtect: A Dynamic Factor Influence Learning Framework for Protective Factor-aware Suicide Risk Prediction},
  author={Li, Jun and Wang, Xiangmeng and Yan, Yifei and Li, Haoyang and Leong, Hong Va and Yu, Nancy Xiaonan and Li, Qing},
  booktitle={Companion Proceedings of the ACM on Web Conference 2025},
  pages={1785--1791},
  year={2025}
}

@article{li2025protective,
  title={Protective Factor-Aware Dynamic Influence Learning for Suicide Risk Prediction on Social Media},
  author={Li, Jun and Wang, Xiangmeng and Li, Haoyang and Yan, Yifei and Leong, Hong Va and Feng, Ling and Yu, Nancy Xiaonan and Li, Qing},
  journal={arXiv preprint arXiv:2507.10008},
  year={2025}
}

@inproceedings{sawhney2021ordinal,
    author={Sawhney, Ramit  and
            Joshi, Harshit  and
            Gandhi, Saumya  and
            Shah, Rajiv Ratn},
    title = {Towards Ordinal Suicide Ideation Detectionon Social Media},
    year = {2021},
    month=mar,
    booktitle = {Proceedings of 14th ACM International Conference On Web Search And Data Mining},
    publisher = {Association for Computing Machinery},
    address = {New York, NY, USA},
    location = {Virtual Event, Israel},
    series = {WSDM '21}
}

@inproceedings{lee2023towards,
  title={Towards suicide prevention from bipolar disorder with temporal symptom-aware multitask learning},
  author={Lee, Daeun and Son, Sejung and Jeon, Hyolim and Kim, Seungbae and Han, Jinyoung},
  booktitle={Proceedings of the 29th ACM SIGKDD conference on knowledge discovery and data mining},
  pages={4357--4369},
  year={2023}
}

@article{kipf2016semi,
  title={Semi-supervised classification with graph convolutional networks},
  author={Kipf, TN},
  journal={arXiv preprint arXiv:1609.02907},
  year={2016}
}

@article{hamilton2017inductive,
  title={Inductive representation learning on large graphs},
  author={Hamilton, Will and Ying, Zhitao and Leskovec, Jure},
  journal={Advances in neural information processing systems},
  volume={30},
  year={2017}
}

@article{velivckovic2017graph,
  title={Graph attention networks},
  author={Veli{\v{c}}kovi{\'c}, Petar and Cucurull, Guillem and Casanova, Arantxa and Romero, Adriana and Lio, Pietro and Bengio, Yoshua},
  journal={arXiv preprint arXiv:1710.10903},
  year={2017}
}

@article{pennebaker2015development,
  title={The development and psychometric properties of LIWC2015},
  author={Pennebaker, James W and Boyd, Ryan L and Jordan, Kayla and Blackburn, Kate},
  year={2015}
}

@incollection{plutchik1980general,
  title={A general psychoevolutionary theory of emotion},
  author={Plutchik, Robert},
  booktitle={Theories of emotion},
  pages={3--33},
  year={1980},
  publisher={Elsevier}
}

@inproceedings{sawhney2021phase,
  title={Phase: Learning emotional phase-aware representations for suicide ideation detection on social media},
  author={Sawhney, Ramit and Joshi, Harshit and Flek, Lucie and Shah, Rajiv},
  booktitle={Proceedings of the 16th conference of the European Chapter of the Association for Computational Linguistics: main volume},
  pages={2415--2428},
  year={2021}
}

@inproceedings{sawhney2020time,
  title={A time-aware transformer based model for suicide ideation detection on social media},
  author={Sawhney, Ramit and Joshi, Harshit and Gandhi, Saumya and Shah, Rajiv},
  booktitle={Proceedings of the 2020 conference on empirical methods in natural language processing (EMNLP)},
  pages={7685--7697},
  year={2020}
}
\end{document}